\pdfoutput=1
\documentclass[11pt,letterpaper]{article}
\usepackage[letterpaper]{geometry}
\usepackage{acl2012} 
\usepackage{times}
\usepackage{latexsym}
\usepackage{microtype}
\usepackage{url}
\usepackage{amsmath}
\usepackage{amsfonts}
\usepackage{multirow}
\usepackage{graphicx}
\usepackage{arydshln}
\usepackage{dsfont}
\usepackage{breqn}
\usepackage{tikz}
\usepackage{amsmath,amsthm,amssymb}
\usepackage{mathtext}
\usepackage[utf8x]{inputenc}
\usepackage[german,bulgarian,portuguese,italian,spanish,dutch,swedish,croatian,polish,russian,english]{babel}
\usepackage[T2A, T1]{fontenc}

\usepackage{amssymb}
\usepackage{pgfplots}

\pgfplotsset{compat=1.12}

\makeatletter
\newcommand{\@BIBLABEL}{\@emptybiblabel}
\newcommand{\@emptybiblabel}[1]{}

\makeatother

\makeatletter

\title{Semantic Specialisation of Distributional Word Vector Spaces using Monolingual and Cross-Lingual Constraints}

\author{Nikola Mrk\v{s}i\'c$^{\mathbf{1, 2}}$, ~ Ivan Vuli\'{c}$^{\mathbf{1}}$, ~ {Diarmuid \'{O} S\'{e}aghdha}$^{\mathbf{2}}$,  ~ \bf{Ira Leviant}$^{\mathbf{3}}$  \\
\bf{Roi Reichart}$^{\mathbf{3}}$, ~ \bf{Milica Ga\v{s}i\'{c}}$^{\mathbf{1}}$,  ~ {Anna Korhonen}$^{\mathbf{1}}$,  ~ {Steve Young}$^{\mathbf{1, 2}}$  \\
$^{\mathbf{1}}$ University of Cambridge \\
$^{\mathbf{2}}$ Apple Inc. \\
$^{\mathbf{3}}$ Technion, IIT}

\begin{document}

\maketitle

\begin{abstract}


We present \textsc{Attract-Repel}, an algorithm for improving the semantic quality of word vectors by injecting constraints extracted from lexical resources. \textsc{Attract-Repel} facilitates the use of constraints from mono- and cross-lingual resources, yielding semantically specialised cross-lingual vector spaces. Our evaluation shows that the method can make use of existing cross-lingual lexicons to construct high-quality vector spaces for a plethora of different languages, facilitating semantic transfer from high- to lower-resource ones. The effectiveness of our approach is demonstrated with state-of-the-art results on semantic similarity datasets in six languages. We next show that \textsc{Attract-Repel}-specialised vectors boost performance in the downstream task of {dialogue state tracking (DST)} across multiple languages. Finally, we show that cross-lingual vector spaces produced by our algorithm facilitate the training of multilingual DST models, which brings further performance improvements.


\end{abstract}


\section{Introduction}

Word representation learning has become a research area of central importance in modern natural language processing. The common techniques for inducing distributed word representations are grounded in the distributional hypothesis, relying on co-occurrence information in large textual corpora to learn meaningful word representations \cite{Mikolov:13,Pennington:14,OSeaghdha:Korhonen:14,Levy:14}. Recently, methods which go beyond stand-alone unsupervised learning have gained increased popularity. These models typically build on distributional ones by using human- or automatically-constructed knowledge bases to enrich the semantic content of existing word vector collections. Often this is done as a post-processing step, where the distributional word vectors are refined to satisfy constraints extracted from a lexical resource such as WordNet \cite{faruqui:15,Wieting:15,Mrksic:16}. We term this approach \textit{semantic specialisation}.


In this paper we advance the semantic specialisation paradigm in a number of ways. We introduce a new algorithm, \textsc{Attract-Repel}, that uses synonymy and antonymy constraints drawn from lexical resources to tune word vector spaces using linguistic information that is difficult to capture with conventional distributional training. Our evaluation shows that \textsc{Attract-Repel} outperforms previous methods which make use of similar lexical resources, achieving state-of-the-art results on two word similarity datasets: SimLex-999 \cite{Hill:2014} and SimVerb-3500 \cite{Gerz:2016}.

\begin{table*}
\def\arraystretch{1.0}
\centering
\resizebox{2.05\columnwidth}{!}{%
{\small
\begin{tabular}{ccc|ccc|ccc}

 \multicolumn{3}{c|}{ \Large \bf en_morning } &  \multicolumn{3}{|c|}{ \Large \bf en_carpet } &  \multicolumn{3}{|c}{ \Large \bf en_woman }  \\ 

 \bf Slavic+EN & \bf Germanic & \bf Romance+EN & \bf Slavic+EN & \bf Germanic & \bf Romance+EN &   \bf Slavic + EN & \bf Germanic & \bf Romance+EN \\ \hline

en\_\foreignlanguage{english}{daybreak} & de\_\foreignlanguage{german}{vormittag} & pt\_\foreignlanguage{portuguese}{madrugada} & en\_\foreignlanguage{english}{rug} & de\_\foreignlanguage{german}{teppichboden} & en\_\foreignlanguage{english}{rug} & ru\_\foreignlanguage{russian}{женщина} & de\_\foreignlanguage{german}{frauen} & fr\_\foreignlanguage{french}{femme} \\ 
en\_\foreignlanguage{english}{morn} & \underline{ nl\_\foreignlanguage{dutch}{krieken} } & it\_\foreignlanguage{italian}{mattina} & bg\_\foreignlanguage{bulgarian}{килим} & nl\_\foreignlanguage{dutch}{tapijten} & it\_\foreignlanguage{italian}{moquette} & bg\_\foreignlanguage{bulgarian}{жените} & sv\_\foreignlanguage{swedish}{kvinnliga} & en\_\foreignlanguage{english}{womanish} \\ 
bg\_\foreignlanguage{bulgarian}{разсъмване} & en\_\foreignlanguage{english}{dawn} & en\_\foreignlanguage{english}{dawn} & ru\_\foreignlanguage{russian}{ковролин} & en\_\foreignlanguage{english}{rug} & it\_\foreignlanguage{italian}{tappeti} & hr\_\foreignlanguage{croatian}{žena} & sv\_\foreignlanguage{swedish}{kvinna} & es\_\foreignlanguage{spanish}{mujer} \\ 
hr\_\foreignlanguage{croatian}{svitanje} & nl\_\foreignlanguage{dutch}{zonsopkomst} & pt\_\foreignlanguage{portuguese}{madrugadas} & bg\_\foreignlanguage{bulgarian}{килими} & de\_\foreignlanguage{german}{teppich} & pt\_\foreignlanguage{portuguese}{tapete} & en\_\foreignlanguage{english}{womanish} & sv\_\foreignlanguage{swedish}{kvinnor} & pt\_\foreignlanguage{portuguese}{mulher} \\ 
hr\_\foreignlanguage{croatian}{zore} & sv\_\foreignlanguage{swedish}{morgonen} & es\_\foreignlanguage{spanish}{madrugada} & pl\_\foreignlanguage{polish}{dywany} & en\_\foreignlanguage{english}{carpeting} & es\_\foreignlanguage{spanish}{moqueta} & bg\_\foreignlanguage{bulgarian}{жена} & de\_\foreignlanguage{german}{weib} & es\_\foreignlanguage{spanish}{fémina} \\ 
bg\_\foreignlanguage{bulgarian}{изгрев} & de\_\foreignlanguage{german}{tagesanbruch} &  \underline{ it\_\foreignlanguage{italian}{nascente} } & bg\_\foreignlanguage{bulgarian}{мокет} & de\_\foreignlanguage{german}{teppiche} & it\_\foreignlanguage{italian}{tappetino} & pl\_\foreignlanguage{polish}{kobieta} & en\_\foreignlanguage{english}{womanish} & en\_\foreignlanguage{english}{womens} \\ 
en\_\foreignlanguage{english}{dawn} & en\_\foreignlanguage{english}{sunrise} & en\_\foreignlanguage{english}{morn} & pl\_\foreignlanguage{polish}{dywanów} & sv\_\foreignlanguage{swedish}{mattor} & en\_\foreignlanguage{english}{carpeting} & hr\_\foreignlanguage{croatian}{treba} & sv\_\foreignlanguage{swedish}{kvinno} & pt\_\foreignlanguage{portuguese}{feminina} \\ 
ru\_\foreignlanguage{russian}{утро} & \underline{ nl\_\foreignlanguage{dutch}{opgang} } & es\_\foreignlanguage{spanish}{aurora} & hr\_\foreignlanguage{croatian}{tepih} & sv\_\foreignlanguage{swedish}{matta} & pt\_\foreignlanguage{portuguese}{carpete} & bg\_\foreignlanguage{bulgarian}{жени} & de\_\foreignlanguage{german}{frauenzimmer} & pt\_\foreignlanguage{portuguese}{femininas} \\ 
 \underline{bg\_\foreignlanguage{bulgarian}{аврора}} & de\_\foreignlanguage{german}{sonnenaufgang} & fr\_\foreignlanguage{french}{matin} & pl\_\foreignlanguage{polish}{wykładziny} & en\_\foreignlanguage{english}{carpets} & pt\_\foreignlanguage{portuguese}{tapetes} & en\_\foreignlanguage{english}{womens} & sv\_\foreignlanguage{swedish}{honkön} & es\_\foreignlanguage{spanish}{femina} \\ 
hr\_\foreignlanguage{croatian}{jutro} & nl\_\foreignlanguage{dutch}{dageraad} &  \underline{ fr\_\foreignlanguage{french}{aurora} } & ru\_\foreignlanguage{russian}{ковер} & nl\_\foreignlanguage{dutch}{tapijt} & fr\_\foreignlanguage{french}{moquette} & pl\_\foreignlanguage{polish}{kobiet} & sv\_\foreignlanguage{swedish}{kvinnan} & fr\_\foreignlanguage{french}{femelle} \\ 
ru\_\foreignlanguage{russian}{рассвет} & de\_\foreignlanguage{german}{anbruch} & es\_\foreignlanguage{spanish}{amaneceres} & ru\_\foreignlanguage{russian}{коврик} & nl\_\foreignlanguage{dutch}{kleedje} & en\_\foreignlanguage{english}{carpets} & hr\_\foreignlanguage{croatian}{žene} & nl\_\foreignlanguage{dutch}{vrouw} & pt\_\foreignlanguage{portuguese}{fêmea} \\ 
hr\_\foreignlanguage{croatian}{zora} & sv\_\foreignlanguage{swedish}{morgon} & en\_\foreignlanguage{english}{sunrises} & hr\_\foreignlanguage{croatian}{ćilim} & nl\_\foreignlanguage{dutch}{vloerbedekking} & es\_\foreignlanguage{spanish}{alfombra} & pl\_\foreignlanguage{polish}{niewiasta} & de\_\foreignlanguage{german}{madam} & fr\_\foreignlanguage{french}{femmes} \\ 
hr\_\foreignlanguage{croatian}{zoru} & en\_\foreignlanguage{english}{daybreak} & es\_\foreignlanguage{spanish}{mañanero} & en\_\foreignlanguage{english}{carpeting} & \underline{de\_\foreignlanguage{german}{brücke}} & es\_\foreignlanguage{spanish}{alfombras} & hr\_\foreignlanguage{croatian}{žensko} & sv\_\foreignlanguage{swedish}{kvinnligt} & it\_\foreignlanguage{italian}{donne} \\ 
pl\_\foreignlanguage{polish}{poranek} & de\_\foreignlanguage{german}{morgengrauen} & fr\_\foreignlanguage{french}{matinée} & pl\_\foreignlanguage{polish}{dywan} & \underline{de\_\foreignlanguage{german}{matta}} & fr\_\foreignlanguage{french}{tapis} & hr\_\foreignlanguage{croatian}{ženke} & { sv\_\foreignlanguage{swedish}{gumman}} & es\_\foreignlanguage{spanish}{mujeres} \\ 
en\_\foreignlanguage{english}{sunrise} & nl\_\foreignlanguage{dutch}{zonsopgang} & it\_\foreignlanguage{italian}{mattinata} & ru\_\foreignlanguage{russian}{ковров} & \underline{nl\_\foreignlanguage{dutch}{matta}} & pt\_\foreignlanguage{portuguese}{tapeçaria} & pl\_\foreignlanguage{polish}{samica} & sv\_\foreignlanguage{swedish}{female} & pt\_\foreignlanguage{portuguese}{fêmeas} \\ 
bg\_\foreignlanguage{bulgarian}{зазоряване} & nl\_\foreignlanguage{dutch}{goedemorgen} & pt\_\foreignlanguage{portuguese}{amanhecer} & en\_\foreignlanguage{english}{carpets} & en\_\foreignlanguage{english}{mat} & it\_\foreignlanguage{italian}{zerbino} & ru\_\foreignlanguage{russian}{самка} & { sv\_\foreignlanguage{swedish}{gumma} } & es\_\foreignlanguage{spanish}{hembras} \\ 
bg\_\foreignlanguage{bulgarian}{сутрин} & sv\_\foreignlanguage{swedish}{gryningen} & en\_\foreignlanguage{english}{cockcrow} & ru\_\foreignlanguage{russian}{килим} & de\_\foreignlanguage{german}{matte} & it\_\foreignlanguage{italian}{tappeto} & bg\_\foreignlanguage{bulgarian}{женска} & sv\_\foreignlanguage{swedish}{kvinnlig} & en\_\foreignlanguage{english}{wife}  \\
en\_\foreignlanguage{english}{sunrises} & en\_\foreignlanguage{english}{mornin} & pt\_\foreignlanguage{portuguese}{aurora} & en\_\foreignlanguage{english}{mat} & en\_\foreignlanguage{english}{doilies} & es\_\foreignlanguage{spanish}{tapete} & hr\_\foreignlanguage{croatian}{ženka} & sv\_\foreignlanguage{swedish}{feminin} & fr\_\foreignlanguage{french}{nana} \\ 
bg\_\foreignlanguage{bulgarian}{зора} & sv\_\foreignlanguage{swedish}{gryning} & pt\_\foreignlanguage{portuguese}{alvorecer} & hr\_\foreignlanguage{croatian}{sag} & nl\_\foreignlanguage{dutch}{mat} & es\_\foreignlanguage{spanish}{manta} & ru\_\foreignlanguage{russian}{дама} & en\_\foreignlanguage{english}{wife} & es\_\foreignlanguage{spanish}{hembra}  
\end{tabular}}%
}
\caption{Nearest neighbours for three example words across Slavic, Germanic and Romance language groups (with English included as part of each word vector collection). Semantically dissimilar words have been underlined. \vspace{-3mm}}

\label{tab:qualitative-table}
\end{table*}

We then deploy the \textsc{Attract-Repel} algorithm in a multilingual setting, using semantic relations extracted from BabelNet \cite{Navigli:12,Ehrmann:14}, a cross-lingual lexical resource, to inject constraints between words of different languages into the word representations. This allows us to embed vector spaces of multiple languages into a single vector space,  exploiting information from high-resource languages to improve the word representations of lower-resource ones. Table~\ref{tab:qualitative-table} illustrates the effects of cross-lingual \textsc{Attract-Repel} specialisation by showing the nearest neighbours for three English words across three cross-lingual spaces. In each case, the vast majority of each words' neighbours are meaningful synonyms/translations.\footnote{Some residual (negative) effects of the distributional hypothesis do persist. For example, \emph{nl_krieken}, which is Dutch for \emph{cherries}, is (presumably) identified as a synonym for \emph{en_morning} due to a song called \emph{`a Morning Wish'} by \emph{Emile Van Krieken}.}


While there is a considerable amount of prior research on joint learning of cross-lingual vector spaces (see Sect.~\ref{sec:crosslingual}), to the best of our knowledge we are the first to apply semantic specialisation to this problem.\footnote{Our approach is not suited for languages for which no lexical resources exist. However, many languages have some coverage in cross-lingual lexicons. For instance, BabelNet 3.7 {automatically} aligns WordNet to Wikipedia, providing accurate cross-lingual mappings between 271 languages. In our evaluation, we demonstrate substantial gains for Hebrew and Croatian, both of which are spoken by less than 10 million people worldwide. }  We demonstrate its efficacy with state-of-the-art results on the four languages in the Multilingual SimLex-999 dataset \cite{Leviant:15}. To show that our approach yields semantically informative vectors for lower-resource languages, we collect intrinsic evaluation datasets for {Hebrew} and {Croatian} and show that cross-lingual specialisation significantly improves word vector quality in these two (comparatively) low-resource languages.

In the second part of the paper, we explore the use of \textsc{Attract-Repel}-specialised vectors in a downstream application. One important motivation for training word vectors is to improve the lexical coverage of supervised models for language understanding tasks, e.g.~question answering \cite{Iyyer:EtAl:14} or textual entailment \cite{Rocktaschel:EtAl:16}. In this work, we use the task of \textit{dialogue state tracking} (DST) for extrinsic evaluation. This task, which arises in the construction of statistical dialogue systems \cite{young:13}, involves understanding the goals expressed by the user and updating the system's distribution over such goals as the conversation progresses and new information becomes available. 

We show that incorporating our specialised vectors into a state-of-the-art neural-network model for DST improves performance on English dialogues. In the multilingual spirit of this paper, we produce new Italian and German DST datasets and show that using \textsc{Attract-Repel}-specialised vectors leads to even stronger gains in these two languages. Finally, we show that our cross-lingual vectors can be used to train a single model that performs DST in all three languages, in each case outperforming the monolingual model. To the best of our knowledge, this is the first work on multilingual training of any component of a statistical dialogue system. Our results indicate that multilingual training holds great promise for bootstrapping language understanding models for other languages, especially for dialogue domains where data collection is very resource-intensive. 

{All resources relating to this paper are available at \url{www.github.com/nmrksic/attract-repel}. These include: \textbf{1)} the  \textsc{Attract-Repel} source code; \textbf{2)} bilingual word vector collections combining English with 51 other languages; \textbf{3)} Hebrew and Croatian intrinsic evaluation datasets; and \textbf{4)} Italian and German Dialogue State Tracking datasets collected for this work.}


\section{Related Work}

\subsection{Semantic Specialisation}

The usefulness of distributional word representations has been demonstrated across many application areas: Part-of-Speech (POS) tagging \cite{Collobert:11}, machine translation \cite{Zou:2013emnlp,Devlin:2014}, dependency and semantic parsing \cite{Socher:2013b,bansal:2014,Chen:2014,Johannsen:2015,Ammar:2016tacl}, sentiment analysis \cite{socher:2013}, named entity recognition \cite{Turian:2010,Guo:2014}, and many others. The importance of semantic specialisation for downstream tasks is relatively unexplored, with improvements in performance so far observed for dialogue state tracking \cite{Mrksic:16,Mrksic:16b}, spoken language understanding \cite{Kim:16,Kim:16b} and judging lexical entailment \cite{Vulic:16b}. 

Semantic specialisation methods (broadly) fall into two categories: \textbf{a)} those which train distributed representations `from scratch' by combining distributional knowledge and lexical information; and \textbf{b)} those which \emph{inject} lexical information into pre-trained collections of word vectors. Methods from both categories make use of similar lexical resources; common examples include WordNet \cite{Miller:95}, FrameNet \cite{Baker:98} or the Paraphrase Databases (PPDB) \cite{ppdb:13,ganitkevitch:14,PPDB2}.

\vspace{1mm}
\textbf{Learning from Scratch:} some methods modify the prior or the regularization of the original training procedure using the set of linguistic constraints \cite{Yu:2014,Xu:2014,Bian:14,kiela:15,Aletras:15}. Other ones modify the skip-gram \cite{Mikolov:13} objective function by introducing semantic constraints \cite{Yih:EtAl:12,Liu:EtAl:15} to train word vectors which emphasise word similarity over relatedness. \newcite{Osborne:16} propose a method for incorporating prior knowledge into the Canonical Correlation Analysis (CCA) method used by Dhillon et al.~\shortcite{Dhillon:2015} to learn spectral word embeddings. While such methods introduce semantic similarity constraints extracted from lexicons, approaches such as the one proposed by Schwartz et al. \shortcite{schwartz-reichart-rappoport:2015:Conll} use \emph{symmetric patterns} \cite{Davidov:2006} to push away antonymous words in their pattern-based vector space. \newcite{Ono:15} combine both approaches, using thesauri and distributional data to train embeddings specialised for capturing antonymy. \newcite{faruqui:2015b} use many different lexicons  to create interpretable {sparse binary vectors} which achieve competitive performance across a range of intrinsic evaluation tasks.

In theory, word representations produced by models which consider distributional and lexical information jointly could be as good (or better) than representations produced by fine-tuning distributional vectors. However, their performance has not surpassed that of fine-tuning methods.\footnote{The SimLex-999 web page (\url{www.cl.cam.ac.uk/~fh295/simlex.html}) lists models with state-of-the-art performance, none of which learn representations jointly.}


\vspace{1mm}
\textbf{Fine-Tuning Pre-trained Vectors:} \newcite{Rothe:2015} fine-tune word vector spaces to improve the representations of synsets/lexemes found in WordNet. \newcite{faruqui:15} and \newcite{Jauhar:2015} use synonymy constraints in a procedure termed \emph{retrofitting} to bring the vectors of semantically similar words close together, while \newcite{Wieting:15} modify the skip-gram objective function to fine-tune word vectors by injecting paraphrasing constraints from PPDB. \newcite{Mrksic:16} build on the retrofitting approach by jointly injecting synonymy and antonymy constraints; the same idea is reassessed by \newcite{Nguyen:2016acl}. \newcite{Kim:16b} further expand this line of work by incorporating semantic intensity information for the constraints, while \newcite{Recski:16} use ensembles of rich \emph{concept dictionaries} to further improve a combined collection of semantically specialised word vectors. 




\textsc{Attract-Repel} is an instance of the second family of models, providing a portable, light-weight approach for incorporating external knowledge into arbitrary vector spaces. In our experiments, we show that \textsc{Attract-Repel} outperforms previously proposed post-processors, setting the new state-of-art performance on the widely used SimLex-999 word similarity dataset. Moreover, we show that starting from distributional vectors allows our method to use existing cross-lingual resources to tie distributional vector spaces of different languages into a unified vector space which benefits from positive semantic transfer between its constituent languages.     




\subsection{Cross-Lingual Word Representations}\label{sec:crosslingual}


Most existing models which induce cross-lingual word representations rely on {cross-lingual distributional information} \cite[inter alia]{Klementiev:2012coling,Zou:2013emnlp,Soyer:2015iclr,Huang:2015emnlp}. These models differ in the cross-lingual signal/supervision they use to tie languages into unified bilingual vector spaces: some models learn on the basis of parallel word-aligned data \cite{Luong:2015nw,Coulmance:2015emnlp} or sentence-aligned data \cite{Hermann:2014iclr,Hermann:2014acl,Chandar:2014nips,Gouws:2015icml}. Other ones require document-aligned data \cite{Sogaard:2015acl,Vulic:2016jair}, while some learn on the basis of available bilingual dictionaries \cite{Mikolov:2013arxiv,Faruqui:2014eacl,Lazaridou:2015acl,Vulic:2016acl,Duong:2016emnlp}. See \newcite{Upadhyay:2016acl} and \newcite{Vulic:2016acl} for an overview of cross-lingual word embedding work. 

The inclusion of cross-lingual information results in shared cross-lingual vector spaces which can: \textbf{a)} boost performance on monolingual tasks such as word similarity \cite{Faruqui:2014eacl,Rastogi:2015naacl,Upadhyay:2016acl}; and \textbf{b)} support cross-lingual tasks such as bilingual lexicon induction \cite{Mikolov:2013arxiv,Gouws:2015icml,Duong:2016emnlp}, cross-lingual information retrieval \cite{Vulic:2015sigir,Mitra:2016arxiv}, and transfer learning for resource-lean languages \cite{Sogaard:2015acl,Guo:2015acl}.


However, prior work on cross-lingual word embedding has tended not to exploit pre-existing linguistic resources such as BabelNet. In this work, we make use of cross-lingual constraints derived from such repositories to induce high-quality cross-lingual vector spaces by facilitating semantic transfer from high- to lower-resource languages. In our experiments, we show that cross-lingual vector spaces produced by \textsc{Attract-Repel} consistently outperform a representative selection of five strong cross-lingual word embedding models 
in both intrinsic and extrinsic evaluation across several languages.

\section{The \textsc{Attract-Repel} Model}


In this section, we propose a new algorithm for producing semantically specialised word vectors by injecting similarity and antonymy constraints into distributional vector spaces. This procedure, which we term \textsc{Attract-Repel}, builds on the Paragram \cite{Wieting:15} and counter-fitting procedures \cite{Mrksic:16}, both of which inject linguistic constraints into existing vector spaces to improve their ability to capture semantic similarity.

Let $V$ be the vocabulary, $S$ the set of synonymous word pairs (e.g.~\emph{intelligent} and \emph{brilliant}), and $A$ the set of antonymous word pairs (e.g.~\emph{vacant} and \emph{occupied}). For ease of notation, let each word pair $(x_l, x_r)$ in these two sets correspond to a vector pair $(\mathbf{x}_l, \mathbf{x}_r)$. The optimisation procedure operates over mini-batches $\mathcal{B}$, where each of these consists of a set of synonymy pairs $\mathcal{B}_{S}$ (of size $k_{1}$) and a set of antonymy pairs $\mathcal{B}_{A}$ (of size $k_2$). Let $T_{S}(\mathcal{B}_{S}) = \lbrack (\mathbf{t}_{l}^{1}, \mathbf{t}_{r}^{1}), \ldots, (\mathbf{t}_{l}^{k_{1}}, \mathbf{t}_{r}^{k_1})\rbrack$ and $T_{A}(\mathcal{B}_{A}) = \lbrack (\mathbf{t}_{l}^1, \mathbf{t}_{r}^1), \ldots, (\mathbf{t}_{l}^{k_2}, \mathbf{t}_{r}^{k_2})\rbrack $ be the pairs of \emph{negative examples} for each synonymy and antonymy example pair. These negative examples are chosen from the $2(k_{1} + k_{2})$ word vectors present in $\mathcal{B}_S \cup \mathcal{B}_A$: 

\begin{itemize}

\item For each synonymy pair $(\mathbf{x}_l, \mathbf{x}_r)$, the negative example pair $(\mathbf{t}_l, \mathbf{t}_r)$ is chosen from the remaining in-batch vectors so that $\mathbf{t}_l$ is the one closest (cosine similarity) to $\mathbf{x}_l$ and $\mathbf{t}_r$ is closest to $\mathbf{x}_r$. 

\item For each antonymy pair $(\mathbf{x}_l, \mathbf{x}_r)$, the negative example pair $(\mathbf{t}_l, \mathbf{t}_r)$ is chosen from the remaining in-batch vectors so that $\mathbf{t}_l$ is the one furthest away from $\mathbf{x}_l$ and $\mathbf{t}_r$ is the one furthest from $\mathbf{x}_r$.

\end{itemize}

\noindent These negative examples are used to: \textbf{a)} force synonymous pairs to be closer to each other than to their respective negative examples; and \textbf{b)} to force antonymous pairs to be further away from each other than from their negative examples. The first term of the cost function pulls synonymous words together:
\begin{eqnarray*} S(\mathcal{B}_S) ~=  \sum_{ (x_l, x_r) \in \mathcal{B}_{S}} & \big[ ~ \tau \left( \delta_{syn} +  \mathbf{x}_l \mathbf{t}_l - \mathbf{x}_l \mathbf{x}_r \right)  \\
+& \tau \left( \delta_{syn} +  \mathbf{x}_r \mathbf{t}_r - \mathbf{x}_l \mathbf{x}_r  \right) \big]
\end{eqnarray*}
\noindent where $\tau(x)=\max(0,x)$ is the hinge loss function and $\delta_{syn}$ is the similarity margin which determines how much closer synonymous vectors should be to each other than to their respective negative examples. The second part of the cost function pushes antonymous word pairs away from each other:
\begin{eqnarray*} A(\mathcal{B}_A) ~= \sum_{(x_l, x_r) \in \mathcal{B}_{A}}  & \big[ ~ \tau\left(\delta_{ant} + \mathbf{x}_l \mathbf{x}_r  - \mathbf{x}_l \mathbf{t}_l \right) \\
+&\tau\left( \delta_{ant} + \mathbf{x}_l \mathbf{x}_r  - \mathbf{x}_r \mathbf{t}_r  \right)  \big] 
\end{eqnarray*}
\noindent In addition to these two terms, we include an additional regularisation term which aims to \emph{preserve} the abundance of high-quality semantic content present in the initial (distributional) vector space, as long as this information does not contradict the injected linguistic constraints. If $V(\mathcal{B})$ is the set of all word vectors present in the given mini-batch, then:
\begin{equation*}
  R(\mathcal{B}_S, \mathcal{B}_A) =  \sum\limits_{ \mathbf{x}_i \in V(\mathcal{B}_S \cup \mathcal{B}_A) }  \lambda_{reg} \left\| \widehat{\mathbf{x}_{i}} - \mathbf{x}_i \right\|_{2} 
 \end{equation*}
\noindent where $\lambda_{reg}$ is the L2 regularisation constant and $\widehat{\mathbf{x}_{i}}$ denotes the original (distributional) word vector for word $x_i$. The final cost function of the \textsc{Attract-Repel} algorithm can then be expressed as:
\begin{equation*}
C(\mathcal{B}_S, \mathcal{B}_A) = S(\mathcal{B}_S) + A(\mathcal{B}_A) + R(\mathcal{B}_S, \mathcal{B}_A)
\end{equation*}

\paragraph{Comparison to Prior Work} \textsc{Attract-Repel} draws inspiration from three methods: \textbf{1)} retrofitting \cite{faruqui:15}; \textbf{2)} \textsc{PARAGRAM} \cite{Wieting:15}; and \textbf{3)} counter-fitting \cite{Mrksic:16}. Whereas retrofitting and \textsc{PARAGRAM} do not consider antonymy, counter-fitting models both {synonymy} and {antonymy}. \textsc{Attract-Repel} differs from this method in two important ways: 



\begin{enumerate}

\item{\bf Context-Sensitive Updates:} Counter-fitting uses \emph{attract} and \emph{repel} terms which pull synonyms together and push antonyms apart without considering their relation to other word vectors. For example, its `attract term' is given by: 

\vspace{2mm}
$ ~~~ Attract(S) = \sum_{(\mathbf{x}_l, \mathbf{x}_r) \in S} \tau( \delta_{syn} - \mathbf{x}_l \mathbf{x}_r) $
\vspace{2mm}

\noindent where $S$ is the set of synonymy constraints and $\delta_{syn}$ is the (minimum) similarity enforced between synonyms.  Conversely, \textsc{Attract-Repel} fine-tunes vector spaces by operating over mini-batches of example pairs, updating word vectors only if the position of their negative example implies a stronger semantic relation than that expressed by the position of its target example. Importantly, \textsc{Attract-Repel} makes fine-grained updates to both the example pair \emph{and} the negative examples, rather than updating the example word pair but ignoring how this affects its relation to all other word vectors.       


\item \textbf{\bf Regularisation:} Counter-fitting preserves {distances} between pairs of word vectors in the initial vector space, trying to `pull' the words' neighbourhoods with them as they move to incorporate external knowledge. The radius of this initial neighbourhood introduces an opaque hyperparameter to the procedure. Conversely, \textsc{Attract-Repel} implements standard L2 regularisation, which `pulls' each vector towards its distributional vector representation. 

\end{enumerate}


In our intrinsic evaluation (Sect. 5), we perform an exhaustive comparison of these  models, showing that \textsc{Attract-Repel} significantly outperforms counter-fitting in both mono- and cross-lingual setups. 







\paragraph{Optimisation} Following \newcite{Wieting:15}, we use the AdaGrad algorithm \cite{Duchi:11} to train the word embeddings for five epochs, which suffices for the {magnitude} of the parameter updates to converge. Similar to \newcite{faruqui:15}, \newcite{Wieting:15} and \newcite{Mrksic:16}, we do not use early stopping. By not relying on language-specific validation sets, the \textsc{Attract-Repel} procedure can induce semantically specialised word vectors for languages with no intrinsic evaluation datasets.\footnote{Many languages are present in semi-automatically constructed lexicons such as BabelNet or PPDB (see the discussion in Sect. 4.2.). However, intrinsic evaluation datasets such as SimLex-999 exist for very few languages, as they require expert translators and skilled annotators.}

\paragraph{Hyperparameter Tuning} We use Spearman's correlation of the final word vectors with the Multilingual WordSim-353 gold-standard {association} dataset \cite{Finkelstein:2002,Leviant:15}. The \textsc{Attract-Repel} procedure has six hyperparameters: the regularization constant $\lambda_{reg}$, the similarity and antonymy margins $\delta_{sim}$ and $\delta_{ant}$, mini-batch sizes $k_1$ and $k_2$, and the size of the PPDB constraint set used for each language (larger sizes include more constraints, but also a larger proportion of false synonyms). We ran a grid search over these for the four SimLex languages, choosing the hyperparameters which achieved the best WordSim-353 score.\footnote{We ran the grid search over $\lambda_{reg} \in \lbrack 10^{-3}, \ldots, 10^{-10} \rbrack$, $\delta_{sim}, \delta_{ant} \in \lbrack 0, 0.1, \ldots, 1.0  \rbrack $, $k_1, k_2 \in [10, 25, 50, 100, 200]$ and over the six PPDB sizes for the four SimLex languages. $\lambda_{reg}=10^{-9}$, $\delta_{sim}=0.6$, $\delta_{ant}=0.0$ and $k_1 = k_2 \in [10, 25, 50]$ consistently achieved the best performance (we use $k_1=k_2=50$ in all experiments for consistency). The PPDB constraint set size \emph{XL} was best for English, German and Italian, and \emph{M} achieved the best performance for Russian.}



\section{Experimental Setup}

\subsection{Distributional Vectors}

We first present our sixteen experimental languages: English ({EN}), German ({DE}), Italian ({IT}), Russian ({RU}), Dutch ({NL}), Swedish ({SV}), French ({FR}), Spanish ({ES}), Portuguese ({PT}), Polish ({PL}), Bulgarian ({BG}), Croatian ({HR}), Irish (GA), Persian (FA) and Vietnamese (VI). The first four languages are those of the Multilingual SimLex-999 dataset. 

For the four SimLex languages, we employ four well-known, high-quality word vector collections: \textbf{a)} The Common Crawl GloVe English vectors from Pennington et al.~\shortcite{Pennington:14}; \textbf{b)} German vectors from Vuli\'c and Korhonen \shortcite{vulic:2016}; \textbf{c)} Italian vectors from Dinu et al.~\shortcite{Dinu:15}; and \textbf{d)} Russian vectors from Kutuzov and Andreev \shortcite{Kutuzov:Andreev:15}.
In addition, for each of the 16 languages we also train the skip-gram with negative sampling variant of the \texttt{word2vec} model \cite{Mikolov:13}, on the latest Wikipedia dump of each language, to induce 300-dimensional word vectors.\footnote{The frequency cut-off was set to 50: words that occurred less frequently were removed from the vocabularies. Other \texttt{word2vec} parameters were set to the standard values \cite{vulic:2016}: $15$ epochs, $15$ negative samples, global (decreasing) learning rate: $0.025$, subsampling rate: $1e-4$.}  

\begin{table}
\centering
\resizebox{1\columnwidth}{!}{%
{\small
\begin{tabular}{c|cc|cc|cc|cc}
 & \multicolumn{2}{|c|}{ \bf English} & \multicolumn{2}{|c|}{ \bf German} & \multicolumn{2}{|c|}{ \bf Italian} & \multicolumn{2}{|c}{ \bf Russian} \\ 
  &  syn &  ant  &  syn &  ant  &  syn &  ant  & syn &  ant \\ \hline
  
 \bf English & \underline{640} & \underline{5} & 246 & 11 & 356 & 24 & 196 & 9\\
 \bf German & - & - & \underline{135} & \underline{2} & 277 & 13 & 175 & 6\\
 \bf Italian & - & - & - & - & \underline{159} & \underline{7} & 220 & 11\\
 \bf Russian & - & - & - & - & - & - & \underline{48} & \underline{1}\\

\end{tabular}}%
}
\caption{Linguistic constraint counts (in thousands). For each language pair, the two figures show the number of injected synonymy and antonymy constraints. Monolingual constraints (the diagonal elements) are underlined. \vspace{-4mm} }

\label{tab:constraint-count}
\end{table}

\subsection{Linguistic Constraints} 

Table \ref{tab:constraint-count} shows the number of monolingual and cross-lingual constraints for the four SimLex languages.

\paragraph{Monolingual Similarity}

We employ the Multilingual Paraphrase Database \cite{ganitkevitch:14}. This resource contains paraphrases automatically extracted from parallel-aligned corpora for ten of our sixteen languages. In our experiments, the remaining six languages (HE, HR, SV, GA, VI, FA) serve as examples of \emph{lower-resource} languages, as they have no monolingual synonymy constraints. 


\paragraph{Cross-Lingual Similarity}

We employ BabelNet, a multilingual semantic network automatically constructed by linking Wikipedia to WordNet \cite{Navigli:12,Ehrmann:14}. BabelNet groups words from different languages into \emph{Babel synsets}. We consider two words from any (distinct) language pair to be synonymous if they belong to (at least) one set of synonymous Babel synsets. We made use of all BabelNet word senses tagged as \emph{conceptual} but ignored the ones tagged as \emph{Named Entities}.



Given a large collection of \emph{cross-lingual} semantic constraints (e.g.~the translation pair \emph{en\_sweet} and \emph{it\_dolce}), \textsc{Attract-Repel} can use them to bring the vector spaces of different languages together into a shared cross-lingual space. Ideally, sharing information across languages should lead to improved semantic content for each language, especially for those with limited monolingual resources.

\paragraph{Antonymy} BabelNet is also used to extract \emph{both} monolingual \emph{and} cross-lingual antonymy constraints. Following \newcite{faruqui:15}, who found PPDB constraints more beneficial than the WordNet ones, we do not use BabelNet for {monolingual} synonymy.

\paragraph{Availability of Resources} Both PPDB and BabelNet are created automatically. However, PPDB relies on large, high-quality parallel corpora such as Europarl \cite{Koehn:2005}. In total, Multilingual PPDB provides collections of paraphrases for 22 languages. On the other hand, BabelNet uses Wikipedia's \emph{inter-language links} and statistical machine translation (Google Translate) to provide cross-lingual mappings for 271 languages. In our evaluation, we show that PPDB and BabelNet can be used jointly to improve word representations for lower-resource languages by tying them into bilingual spaces with high-resource ones. We validate this claim on Hebrew and Croatian, which act as `lower-resource' languages because of their lack of any PPDB resource and their relatively small Wikipedia sizes.\footnote{Hebrew and Croatian Wikipedias (which are used to induce their BabelNet constraints) currently consist of 203,867 / 172,824 articles, ranking them 40th / 42nd by size.} 

\section{Intrinsic Evaluation}

\subsection{Datasets}

Spearman's rank correlation with the SimLex-999 dataset \cite{Hill:2014} is used as the intrinsic evaluation metric throughout the experiments. Unlike other gold standard resources such as WordSim-353 \cite{Finkelstein:2002} or MEN \cite{Bruni:2014jair}, SimLex-999 consists of word pairs scored by annotators instructed to discern between semantic similarity and conceptual association, so that related but non-similar words (e.g.~\emph{book} and \emph{read}) have a low rating. 

\newcite{Leviant:15} translated SimLex-999 to German, Italian and Russian, crowd-sourcing the similarity scores from native speakers of these languages. We use this resource for  multilingual intrinsic  evaluation.\footnote{\newcite{Leviant:15} also re-scored the original English SimLex. We report results on their version, but also provide numbers for the original dataset for comparability.} To investigate the portability of our approach to lower-resource languages, we used the same experimental setup to collect SimLex-999 datasets for Hebrew and Croatian.\footnote{The 999 word pairs and annotator instructions were translated by native speakers and scored by $10$ annotators. The inter-annotator agreement scores (Spearman's $\rho$) were 0.77 (pairwise) and 0.87 (mean) for Croatian, and 0.59 / 0.71 for Hebrew.} 
For English vectors, we also report Spearman's correlation with SimVerb-3500 \cite{Gerz:2016}, a semantic similarity dataset that focuses on verb pair similarity. 

\begin{table*} [!h]
\def\arraystretch{0.95}
\centering
{\small
\resizebox{2.0\columnwidth}{!}{%
\begin{tabular}{l|cccc}
\bf Word Vectors & \bf English & \bf German & \bf Italian & \bf Russian \\ \hline
 \bf Monolingual Distributional Vectors & 0.32 & 0.28 & 0.36 & 0.38 \\ \hline 
 
\textsc{Counter-Fitting}: Mono-Syn & 0.45 & 0.24 & 0.29 & 0.46 \\
\textsc{Counter-Fitting}: Mono-Ant & 0.33 & 0.28  & 0.47 & 0.42 \\
\textsc{Counter-Fitting}: Mono-Syn + Mono-Ant & 0.50 & 0.26 & 0.35 & 0.49 \\ \hdashline
\textsc{Counter-Fitting}: Cross-Syn & 0.46 & 0.43 & 0.45 & 0.37 \\
\textsc{Counter-Fitting}: Mono-Syn + Cross-Syn & 0.47 & 0.40 & 0.43 & 0.45 \\
\textsc{Counter-Fitting}: Mono-Syn + Mono-Ant + Cross-Syn + Cross-Ant & 0.53 & 0.41  & 0.49 & 0.48 \\ \hline

\textsc{Attract-Repel}: Mono-Syn & 0.56 & 0.40 & 0.46 & 0.53 \\
\textsc{Attract-Repel}: Mono-Ant & 0.42  & 0.30 & 0.45 & 0.41 \\
\textsc{Attract-Repel}: Mono-Syn + Mono-Ant & 0.65 & 0.43 & 0.56 & 0.56 \\ \hdashline
\textsc{Attract-Repel}: Cross-Syn & 0.57 & 0.53 & 0.58 & 0.46 \\
\textsc{Attract-Repel}: Mono-Syn + Cross-Syn & 0.61 & 0.58  & 0.59 & 0.54 \\
\textsc{Attract-Repel}: Mono-Syn + Mono-Ant + Cross-Syn + Cross-Ant & \bf 0.71 & \bf 0.62 & \bf 0.67 & \bf 0.61 
\end{tabular}}

\caption{Multilingual SimLex-999. The effect of using the \textsc{Counter-Fitting} and \textsc{Attract-Repel} procedures to inject mono- and cross-lingual synonymy and antonymy constraints into the four collections of distributional word vectors. Our best results set the new state-of-the-art performance for all four languages. \vspace{-0mm}}  
\label{tab:simlex-experiments}}
\end{table*}

\subsection{Experiments}

\paragraph{Monolingual and Cross-Lingual Specialisation} We start from distributional vectors for the SimLex languages: English, German, Italian and Russian. For each language, we first perform semantic specialisation of these spaces using: a) monolingual synonyms; b) monolingual antonyms; and c) the combination of both. We then add cross-lingual synonyms and antonyms to these constraints and train a shared four-lingual vector space for these languages.     



\paragraph{Comparison to Baseline Methods} Both mono- and cross-lingual specialisation was performed using \textsc{Attract-Repel} {and} counter-fitting, in order to conclusively determine which of the two methods exhibited superior performance. Retrofitting and \textsc{PARAGRAM} methods only inject synonymy, and their cost functions can be expressed using sub-components of counter-fitting and \textsc{Attract-Repel} cost functions. As such, the performance of the two investigated methods when they make use of similarity (but not antonymy) constraints illustrates the performance range of the two preceding models.  

\paragraph{Importance of Initial Vectors} We use three different sets of initial vectors: \textbf{a)} well-known distributional word vector collections (Sect. 4.1); \textbf{b)} distributional vectors trained on the latest Wikipedia dumps; and \textbf{c)} word vectors randomly initialised using the \textsc{xavier} initialisation \cite{Glorot:2010aistats}. 


\paragraph{Specialisation for Lower-Resource Languages} In this experiment, we first construct bilingual spaces which combine: \textbf{a)} one of the four SimLex languages; with \textbf{b)} each of the other twelve languages.\footnote{Hyperparameters: we used $\delta_{sim}=0.6$, $\delta_{ant}=0.0$ and  $\lambda_{reg}=10^{-9}$, which achieved the best performance when tuned for the original SimLex languages. The largest available PPDB size was used for the six languages with available PPDB (French, Spanish, Portuguese, Polish, Bulgarian and Dutch).} Since each pair contains at least one SimLex language, we can analyse the improvement over  {monolingual specialisation} to understand how robust the performance gains are across different language pairs. We next use the newly collected SimLex datasets for Hebrew and Croatian to evaluate the extent to which bilingual semantic specialisation using \textsc{Attract-Repel} and BabelNet constraints can improve word representations for lower-resource languages.




\paragraph{Comparison to State-of-the-Art Bilingual Spaces} The English-Italian and English-German bilingual spaces induced by \textsc{Attract-Repel} were compared to five state-of-the-art methods for constructing bilingual vector spaces: \textbf{1.} \cite{Mikolov:2013arxiv}, re-trained using the constraints used by our model; and \textbf{2.-5.} \cite{Hermann:2014iclr,Gouws:2015icml,vulic:2016,Vulic:2016jair}. The latter models use various sources of supervision (word-, sentence- and document-aligned corpora), which means they cannot be trained using our sets of constraints. For these models, we use competitive setups proposed in \cite{vulic:2016}. The goal of this experiment is to show that vector spaces induced by \textsc{Attract-Repel} exhibit better intrinsic and extrinsic performance when deployed in language understanding tasks. 


\begin{table} 
\centering
\resizebox{1.0\linewidth}{!}{%
\small{
\begin{tabular}{c|cccc}
\bf Word Vectors & \bf EN & \bf DE & \bf IT & \bf RU \\ \hline
\bf Random Init. (No Info.) & 0.01 & -0.03 & 0.02 & -0.03 \\ 
\bf \textsc{A-R}: Monolingual Cons. & 0.54  & 0.33 & 0.29 & 0.35 \\
\bf \textsc{A-R}: Mono + Cross-Ling. &  0.66 &  0.49 &  0.59 &  0.51 \\ \hline
\bf Distributional Wiki Vectors & 0.32 & 0.31 & 0.28 & 0.19 \\ 
\bf \textsc{A-R}: Monolingual Cons. & 0.61 & 0.48 & 0.53 & 0.52 \\
\bf \textsc{A-R}: Mono + Cross-Ling. &  0.66 &  0.60 &  0.65 &  0.54 \\ 
\end{tabular}}}%
\caption{
Multilingual SimLex-999. The effect of \textsc{Attract-Repel} (A-R) on alternative sets of starting word vectors (Random = \textsc{xavier} initialisation).
\vspace{-0mm}}

\label{tab:simlex-experiments-wiki}
\end{table}

\subsection{Results and Discussion} 

Table \ref{tab:simlex-experiments} shows the effects of monolingual and cross-lingual semantic specialisation of four well-known distributional vector spaces for the SimLex languages. Monolingual specialisation leads to very strong improvements in the SimLex performance across all languages. Cross-lingual specialisation brings further improvements, with all languages benefiting from sharing the cross-lingual vector space. Italian in particular shows strong evidence of effective transfer, with Italian vectors' performance coming close to the top-performing English ones.

\paragraph{Comparison to Baselines} Table \ref{tab:simlex-experiments} gives an exhaustive comparison of \textsc{Attract-Repel} to counter-fitting: \textsc{Attract-Repel} achieved substantially stronger performance in all experiments. We believe these results conclusively show that the {fine-grained updates} and {L2 regularisation} employed by \textsc{Attract-Repel} present a better alternative to the context-insensitive attract/repel terms and pair-wise regularisation employed by counter-fitting.

\paragraph{State-of-the-Art} \newcite{Wieting:2016} note that the hyperparameters of the widely used Paragram-SL999 vectors \cite{Wieting:15} are tuned on SimLex-999, and as such are not comparable to methods which holdout the dataset. This implies that further work which uses these vectors (e.g.,~ \cite{Mrksic:16,Recski:16}) as starting point does not yield meaningful high scores either. Our reported English score of 0.71 on the Multilingual SimLex-999 corresponds to \textbf{0.751} on the original SimLex-999: it outperforms the 0.706 score reported by \newcite{Wieting:2016} and sets a new high score for this dataset. Similarly, the SimVerb-3500 score of these vectors is \textbf{0.674}, outperforming the current state-of-the-art score of 0.628 reported by \newcite{Gerz:2016}.

\begin{table*} [!h]
\def\arraystretch{0.95}
\centering
\resizebox{2.0\columnwidth}{!}{%
{\small
\begin{tabular}{c|c|cccc|cccccc|ccccccc}
& \bf Mono. & \multicolumn{4}{c|}{ SimLex Languages} &  \multicolumn{6}{c|}{ PPDB available} &  \multicolumn{6}{c}{ No PPDB available} \\
& \bf Spec. &   \bf EN & \bf DE & \bf IT & \bf RU & \bf NL  & \bf FR & \bf ES & \bf   PT & \bf PL & \bf BG & \bf HR  & \bf HE & \bf GA & \bf VI & \bf FA  & \bf SV \\ \hline
\bf  English  & {0.65} & - &  0.69    &    0.70     &    0.70 &    0.70 &      0.72   &    0.72  &   0.70 &    0.70 &   0.68 &   0.70 &    0.66 &   0.65 &   0.67 &   0.68 &   0.70  \\
\bf  German    & 0.43  &    0.61 & - &    0.58 &   0.56  &   0.55 &   0.60  &   0.59 &     0.56  &   0.54 &    0.52 &    0.53 &   0.50 &   0.49 &   0.48 &   0.51 & 0.55 \\
\bf   Italian   & 0.56  &  0.69 &   0.65 & - &   0.64   &    0.67   &   0.68 &   0.68 &   0.66 &   0.66 &    0.62 &   0.63 &   0.59 &    0.60 &   0.58 &   0.61 &   0.63 \\
\bf   Russian   & 0.56 &    0.63 &    0.59  &   0.62  & - &   0.61 &   0.61  &   0.62  &   0.58  &   0.60 &   0.61 &   0.59   &  0.56 &    0.57 &   0.58 &   0.58 &   0.60
 \end{tabular}}%
}
\caption{SimLex-999 performance. Tying the SimLex languages into bilingual vector spaces with 16 different languages. The first number in each row represents monolingual specialisation. {All} but two of the bilingual spaces improved over these baselines. The EN-FR vectors set a new high score of \textbf{0.754} on the original (English) SimLex-999. \vspace{-0mm} }

\label{tab:four-lang-expansion}
\end{table*}

\begin{table} 
\def\arraystretch{0.9}
\centering
{\small
\begin{tabular}{c|c|cccc}
     & \bf Distrib. & \bf  + EN  & \bf  + DE   & \bf + IT  & \bf + RU \\ \hline
\bf Hebrew    & 0.28 & \bf 0.51 &  0.46 & 0.52  & 0.45  \\
\bf Croatian   & 0.21 & \bf 0.62  & 0.49  & 0.58  & 0.54    \\ \hline
\bf English   & 0.32 & - &  0.61 & \bf 0.66 & 0.63 \\
\bf German   & 0.28 &  \bf 0.58 & - &  0.55 &  0.49 \\ 
\bf Italian   & 0.36 & \bf 0.69 &  0.66 & - &  0.63 \\
\bf Russian   & 0.38 & \bf 0.56 & 0.52 & 0.55 & - \\  











\end{tabular}
}
\caption{Bilingual semantic specialisation for: \textbf{a)} Hebrew and Croatian; and \textbf{b)} the original SimLex languages. Each row shows how SimLex scores for that language improve when its distributional vectors are tied into bilingual vector spaces with the four high-resource languages. \vspace{-0mm} } 

\label{tab:low-res-simulation}
\end{table}

\paragraph{Starting Distributional Spaces} Table~\ref{tab:simlex-experiments-wiki} repeats the previous experiment with two different sets of initial vector spaces: \textbf{a)} randomly initialised word vectors;\footnote{The \textsc{xavier} initialisation populates the values for each word vector by uniformly sampling from the interval $[-\frac{\sqrt{6}}{\sqrt{d}},+\frac{\sqrt{6}}{\sqrt{d}}]$, where $d$ is the vector dimensionality. This is a typical init method in neural nets research \cite{Goldberg:2015primer,Bengio:2013pami}.} and \textbf{b)} skip-gram with negative sampling vectors trained on the latest Wikipedia dumps. The randomly initialised vectors serve to decouple the impact of injecting external knowledge from the information embedded in the distributional vectors. The random vectors benefit from both mono- and cross-lingual specialisation: the English performance is surprisingly strong, with other languages suffering more from the lack of initialisation.

When comparing distributional vectors trained on Wikipedia to the high-quality word vector collections used in Table~\ref{tab:simlex-experiments}, the Italian and Russian vectors in particular start from substantially weaker SimLex scores. The difference in performance is largely mitigated through semantic specialisation. However, all vector spaces still exhibit weaker performance compared to those in Table \ref{tab:simlex-experiments}. We believe this shows that the quality of the initial distributional vector spaces is important, but can in large part be compensated for through semantic specialisation.

\paragraph{Bilingual Specialisation} Table \ref{tab:four-lang-expansion} shows the effect of combining the four original SimLex languages with each other and with twelve other languages (Sect. 4.1). Bilingual specialisation substantially improves over monolingual specialisation for \emph{all language pairs}. This indicates that our improvements are language independent to a large extent. 

Interestingly, even though we use no monolingual synonymy constraints for the six right-most languages, combining them with the SimLex languages still improved word vector quality for these four high-resource languages. The reason why even resource-deprived languages such as Irish help improve vector space quality of high-resource ones such as English or Italian is that they provide {implicit} indicators of semantic similarity. English words which map to the same Irish word are likely to be synonyms, even if those English pairs are not present in the PPDB datasets \cite{Faruqui:2014eacl}.\footnote{We release bilingual vector spaces for \textsc{EN} + 51 other languages: the 16 presented here and another 35 languages (all available at \url{www.github.com/nmrksic/attract-repel}). }   

\paragraph{Lower-Resource Languages} The previous experiment indicates that bilingual specialisation further improves the (already) high-quality estimates for high-resource languages. However, it does little to show how much (or if) the word vectors of lower-resource languages improve during such specialisation. Table \ref{tab:low-res-simulation} investigates this proposition using the newly collected SimLex datasets for Hebrew and Croatian. 

Tying the distributional vectors for these languages (which have no monolingual constraints) into cross-lingual spaces with high-resource ones (which do, in our case from PPDB) leads to substantial improvements. Table \ref{tab:low-res-simulation} also shows how the distributional vectors of the four SimLex languages improve when tied to other languages (in each row, we use monolingual constraints only for the `added' language). Hebrew and Croatian exhibit similar trends to the original SimLex languages: tying to English and Italian leads to stronger gains than tying to the morphologically sophisticated German and Russian. Indeed, tying to English consistently lead to strongest performance. We believe this shows that bilingual \textsc{Attract-Repel} specialisation with English promises to produce high-quality vector spaces for many lower-resource languages which have coverage among the 271 BabelNet languages (but are not available in PPDB). 

\paragraph{Existing Bilingual Spaces} Table~\ref{tab:baselines} compares the intrinsic (i.e.~SimLex-999) performance of bilingual English-Italian and English-German vectors produced by \textsc{Attract-Repel} to five previously proposed approaches for constructing bilingual vector spaces. For both languages in both language pairs, \textsc{Attract-Repel} achieves substantial gains over all of these methods.  In the next section, we show that these differences in intrinsic performance lead to substantial gains in downstream evaluation.   

\begin{table} 
\resizebox{1.0\columnwidth}{!}{%
{\small
\begin{tabular}{c|rc|rc}
\multirow{2}{*}{\bf Model} & \multicolumn{2}{c|}{\bf EN-IT} & \multicolumn{2}{c}{\bf EN-DE} \\ 
 & \bf EN & \bf  IT & \bf EN & \bf  DE \\ \hline
\bf \cite{Mikolov:2013arxiv} & 0.32 & 0.28 & 0.32 & 0.28 \\ 
\bf \cite{Hermann:2014iclr} & 0.40 & 0.34 & 0.38 & 0.35 \\ 
\bf \cite{Gouws:2015icml} & 0.25 & 0.18 & 0.25 & 0.14  \\
\bf \cite{vulic:2016} & 0.32 & 0.27 & 0.32 & 0.33 \\
\bf \cite{Vulic:2016jair} & 0.23 & 0.25 & 0.20 & 0.25 \\ \hline
\bf Bilingual \textsc{Attract-Repel} & \bf 0.70 & \bf 0.69 &  \bf 0.69 & \bf 0.61  
\end{tabular}}%
}
\caption{Comparison of the intrinsic quality (SimLex-999) of bilingual spaces produced by the \textsc{Attract-Repel} method to those produced by five state-of-the-art methods for constructing bilingual vector spaces. 
\label{tab:baselines} \vspace{-0mm}} 
\end{table}



\section{Downstream Task Evaluation}

\subsection{Dialogue State Tracking}

Task-oriented dialogue systems help users achieve goals such as making travel reservations or finding restaurants. In \emph{slot-based} systems, application domains are defined by \emph{ontologies} which enumerate the goals that users can express \cite{young:10b}. The goals are expressed by \emph{slot-value} pairs such as [price: \emph{cheap}] or [food: \emph{Thai}]. For modular task-based systems, the Dialogue State Tracking (DST) component is in charge of maintaining the \emph{belief state}, which is the system's internal distribution over the possible states of the dialogue. Figure \ref{fig:example-dialogue} shows the correct dialogue state for each turn of an example dialogue.

\paragraph{Unseen Data/Labels} As dialogue ontologies can be very large, many of the possible \emph{class labels} (i.e.,~the various \emph{food types} or \emph{street names}) will not occur in the training set. To overcome this problem, \emph{delexicalisation-based} DST models \cite{Henderson:14b,Henderson:14d,Mrksic:15,Wen:16} replace occurrences of ontology values with generic tags which facilitate transfer learning across different ontology values. This is done through exact matching supplemented with \emph{semantic lexicons} which encode rephrasings, morphology and other linguistic variation. For instance, such lexicons would be required to deal with the underlined non-exact matches in Figure \ref{fig:example-dialogue}.  


\paragraph{Exact Matching as a Bottleneck} Semantic lexicons can be hand-crafted for small dialogue domains. \newcite{Mrksic:16} showed that semantically specialised vector spaces can be used to automatically induce such lexicons for simple dialogue domains. However, as domains grow more sophisticated, the reliance on (manually- or automatically-constructed) semantic dictionaries  which list potential rephrasings for ontology values becomes a bottleneck for deploying dialogue systems. Ambiguous rephrasings are just one problematic instance of this approach: a user asking about \emph{Iceland} could be referring to the country or the supermarket chain, and someone asking for songs by \emph{Train} is not interested in train timetables. More importantly, the use of English as the principal language in most dialogue systems research understates the challenges that complex linguistic phenomena present in other languages. In this work, we investigate the extent to which semantic specialisation can empower DST models which \emph{do not rely} on such dictionaries.  


\begin{figure} 
\begin{tabular}{p{7cm}}
  \textbf{User:} Suggest something \underline{fancy}.\\
  \texttt{[price=\underline{expensive}]}\\[0.4ex]
  \textbf{System:} Sure, where?\\
  \textbf{User:} \underline{Downtown}. Any {Korean} places? \\
  \texttt{[price=expensive, area=\underline{centre}, ~~food={Korean}]}\\[0.4ex]
  \textbf{System:} Sorry, no Korean places in the centre.\\
  \textbf{User:} How about {Japanese}? \\
  \texttt{[price=expensive, area=centre, ~~food={Japanese}]}\\[0.4ex]
  \textbf{System:} Sticks'n'Sushi meets your criteria. 
\end{tabular} 
\caption{Annotated dialogue states in a sample dialogue. Underlined words show rephrasings for ontology values which are typically handled using semantic dictionaries. \vspace{-0mm} \label{fig:example-dialogue} }
\end{figure}

\paragraph{Neural Belief Tracker (NBT)} The NBT is a novel DST model which operates purely over distributed representations of words, learning to compose utterance and context representations which it then uses to decide which of the potentially many ontology-defined intents (goals) have been expressed by the user \cite{Mrksic:16b}. To overcome the data sparsity problem, the NBT uses \emph{label embedding} to decompose this multi-class classification problem into many binary classification ones: for each slot, the model iterates over slot values defined by the ontology, deciding whether each of them was expressed in the current utterance and its surrounding context.  The first NBT layer consists of neural networks which produce distributed representations of the user utterance,\footnote{There are two variants of the NBT model: \textbf{NBT-DNN} and \textbf{NBT-CNN}. In this work, we limit our investigation to the latter one, as it achieved consistently stronger DST performance.} the preceding system output and the \emph{embedded label} of the candidate slot-value pair. These representations are then passed to the downstream \emph{semantic decoding} and \emph{context modelling} networks, which subsequently make the binary decision regarding the current slot-value candidate. When contradicting goals are detected (i.e.~\emph{cheap} and \emph{expensive}), the model chooses the more probable one.

The NBT training procedure keeps the initial word vectors fixed: that way, at test time, unseen words semantically related to familiar slot values (i.e.~\emph{affordable} or \emph{cheaper} to \emph{cheap}) are recognised purely by their position in the original vector space. Thus, it is essential that deployed word vectors are specialised for semantic similarity, as distributional effects which keep antonymous words' vectors together can be very detrimental to DST performance (e.g., by matching \emph{northern} to \emph{south} or \emph{inexpensive} to \emph{expensive}).

\paragraph{The Multilingual WOZ 2.0 Dataset} Our DST evaluation is based on the WOZ 2.0 dataset introduced by \newcite{Wen:16} and \newcite{Mrksic:16b}. This dataset is based on the ontology used for the 2nd DST Challenge (DSTC2) \cite{Henderson:14a}. It consists of 1,200 Wizard-of-Oz \cite{Fraser:Gilbert:91} dialogues in which Amazon Mechanical Turk users assumed the role of the dialogue system or the caller looking for restaurants in Cambridge, UK. Since users typed instead of using speech and interacted with intelligent assistants, the language they used was more sophisticated than in case of DSTC2, where users would quickly adapt to the system's inability to cope with complex queries. For our experiments, the ontology and 1,200 dialogues were translated to Italian and German through \url{gengo.com}, a web-based human translation platform. 

\subsection{DST Experiments}

The principal evaluation metric in our DST experiments is the \emph{joint goal accuracy}, which represents the proportion of test set dialogue turns where all the search constraints expressed up to that point in the conversation were decoded correctly. Our DST experiments investigate two propositions:

\begin{enumerate}
\item{\bf Intrinsic vs. Downstream Evaluation} If mono- and cross-lingual semantic specialisation improves the semantic content of word vector collections according to intrinsic evaluation, we would expect the NBT model to perform higher-quality belief tracking when such improved vectors are deployed. We investigate the difference in DST performance for English, German and Italian when the NBT model employs the following word vector collections: \textbf{1)} distributional word vectors; \textbf{2)} monolingual semantically specialised vectors; and \textbf{3)} monolingual subspaces of the cross-lingual semantically specialised EN-DE-IT-RU vectors. For each language, we also compare to the NBT performance achieved using the five state-of-the-art bilingual vector spaces we compared to in Sect. 5.3. 


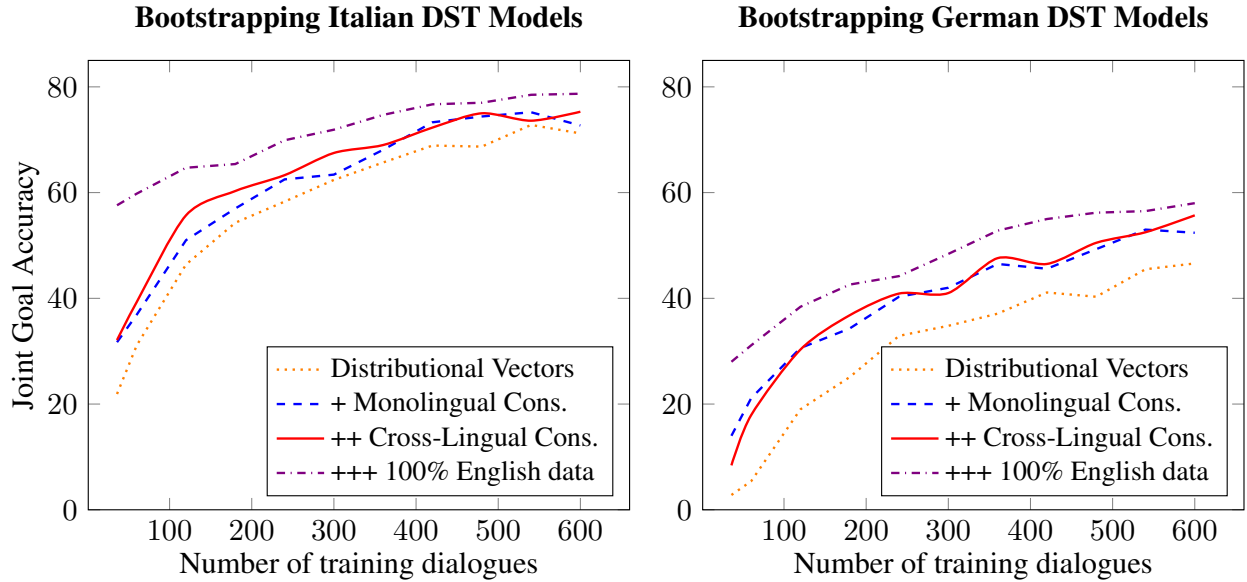
\begin{figure*} [ht]
\centering
\resizebox{2.1\columnwidth}{!}{%
\begin{minipage}{1.0\columnwidth}
\centering
\begin{tikzpicture}
\begin{axis}[
x tick label style={ /pgf/number format/1000 sep=},
legend pos=south east,
legend cell align=left,
scaled y ticks=true,
xmin=1,ymin=0,ymax=85.0,
x label style={at={(current axis.south)},anchor=north, below=4mm},
y label style={at={(current axis.west)},rotate=0,anchor=north, below=-11mm},
xlabel={Number of training dialogues},
ylabel={Joint Goal Accuracy},
title={\bf Bootstrapping Italian DST Models},
legend entries={\small{Distributional Vectors}, \small{+ Monolingual Cons.}, \small{++ Cross-Lingual Cons.}, \small{+++ 100\% English data }}
]
\addplot[orange,dotted,line width=0.3mm] table {italiandist.txt};
\addplot[blue,dashed,line width=0.3mm] table {italiansm.txt};
\addplot[red,smooth,line width=0.3mm] table {italiancrossling.txt};
\addplot[violet,dashdotted,line width=0.3mm] table {italian-grounding.tex};

\end{axis}

\end{tikzpicture}\end{minipage}\hfill
\begin{minipage}{1.0\columnwidth}
\centering
\begin{tikzpicture}
\begin{axis}[
x tick label style={ /pgf/number format/1000 sep=},
legend pos=south east,
legend cell align=left,
scaled y ticks=true,
xmin=1,ymin=0,ymax=85.0,
x label style={at={(current axis.south)},anchor=north, below=4mm},
y label style={at={(current axis.west)},rotate=0,anchor=north, below=-11mm},
xlabel={Number of training dialogues},
title={\bf Bootstrapping German DST Models},
legend entries={\small{Distributional Vectors}, \small{+ Monolingual Cons.}, \small{++ Cross-Lingual Cons.},  \small{+++ 100\% English data}}
]
\addplot[orange,dotted,line width=0.3mm] table {germandist.txt};
\addplot[blue,dashed,line width=0.3mm] table {germansm.txt};
\addplot[red,smooth,line width=0.3mm] table {germancrossling.txt};
\addplot[violet,dashdotted,line width=0.3mm] table {german-grounding.tex};

\end{axis}
\end{tikzpicture}
\end{minipage}}
\caption{Joint goal accuracy of the NBT-CNN model for Italian (left) and German (right) WOZ 2.0 test sets as a function of the number of in-language dialogues used for training.  \vspace{-0mm} \label{fig:dst-curve} }
\end{figure*}

\item \textbf{Training a Multilingual DST Model} The values expressed by the domain ontology (e.g., \emph{cheap}, \emph{north}, \emph{Thai}, etc.)  are language independent. If we assume common semantic grounding across languages, we can \emph{decouple} the ontologies from the dialogue corpora and use a \emph{single ontology} (i.e.~its values' vector representations) across all languages. Since we know that high-performing English DST is attainable, we will \emph{ground} the Italian and German ontologies (i.e.~all slot-value pairs) to the original English ontology. The use of a single ontology coupled with cross-lingual vectors then allows us to combine the training data for multiple languages and train a single NBT model capable of performing belief tracking across all three languages at once. Given a high-quality cross-lingual vector space, combining the languages effectively increases the training set size and should therefore lead to improved performance across all languages.  

\end{enumerate}

\subsection{Results and Discussion}

The DST performance of the NBT-CNN model on English, German and Italian WOZ 2.0 datasets is shown in Table \ref{tab:dst_performance}. The first five rows show the performance when the model employs the five baseline vector spaces. The subsequent three rows show the performance of: \textbf{a)} distributional vector spaces;  \textbf{b)} their monolingual specialisation; and \textbf{c)} their EN-DE-IT-RU cross-lingual specialisation. The last row shows the performance of the {multilingual} DST model trained using \emph{ontology grounding}, where the training data of all three languages was combined and used to train an improved model. Figure \ref{fig:dst-curve} investigates the usefulness of ontology grounding for bootstrapping DST models for new languages with less data: the two figures display the Italian / German performance of models trained using different proportions of the in-language training dataset. The top-performing dash-dotted curve shows the performance of the model trained using the language-specific dialogues and all of the English training data.

The results in Table \ref{tab:dst_performance} show that both types of specialisation improve over DST performance achieved using the distributional vectors or the five baseline bilingual spaces. Interestingly, the bilingual vectors of \newcite{vulic:2016} outperform ours for EN (but not for IT and DE) despite their weaker SimLex performance, showing that intrinsic evaluation does not capture all relevant aspects pertaining to word vectors' usability for downstream tasks. 

The multilingual DST model trained using ontology grounding offers substantial performance improvements, with particularly large gains in the low-data scenario investigated in Figure \ref{fig:dst-curve} (dash-dotted purple line). This figure also shows that the difference in performance between our mono- and cross-lingual vectors is not very substantial. Again, the large disparity in SimLex scores induced only minor improvements in DST performance. 

\begin{table} 
\centering
\resizebox{1.0 \columnwidth}{!}{%
{\small
\begin{tabular}{l|c|c|c}
\multicolumn{1}{c|}{\multirow{1}{*}{\bf Word Vector Space}} &  \bf EN &  {\bf IT} & {\bf DE } \\  \hline
\bf EN-IT/EN-DE \cite{Mikolov:2013arxiv} & 78.2 & 71.1 & 50.5  \\
\bf  EN-IT/EN-DE (Hermann et al.,~2014a) & 71.7  & 69.3 & 44.7 \\ 
\bf  EN-IT/EN-DE \cite{Gouws:2015icml}  & 75.0 & 68.4 &  45.4    \\
\bf  EN-IT/EN-DE (Vuli\'{c} et al.,~2016a)  &  81.6 &  71.8 &  50.5  \\
\bf  EN-IT/EN-DE (Vuli\'{c} et al.,~2016) &  72.3 & 69.0  & 38.2 \\ \hline 
\bf  Monolingual Distributional Vectors & 77.6 &  71.2    & 46.6   \\ 
\bf  \textsc{A-R}: Monolingual Specialisation & 80.9 &  72.7    &  52.4  \\  
\bf  \textsc{A-R}: Cross-Lingual Specialisation & 80.3 &  75.3 &  55.7 \\ \hline
\bf  + English Ontology Grounding & \bf 82.8 & \bf 77.1 &  \bf 57.7  \\
\end{tabular}
}}
\caption{NBT model accuracy across the three languages. Each figure shows the performance of the model trained using the subspace of the given vector space corresponding to the target language. For the English baseline figures, we show the stronger of the EN-IT / EN-DE figures. 
\label{tab:dst_performance} \vspace{-2mm}} 
\end{table}

In summary, our results show that: \textbf{a)} semantically specialised vectors benefit DST performance; \textbf{b)} large gains in SimLex scores do not always induce large downstream gains; and \textbf{c)} high-quality cross-lingual spaces facilitate transfer learning between languages and offer an effective method for bootstrapping DST models for lower-resource languages. 

Finally, German DST performance is substantially weaker than both English and Italian, corroborating our intuition that linguistic phenomena such as cases and compounding make German DST very challenging. We release these datasets in hope that multilingual DST evaluation can give the NLP community a tool for evaluating downstream performance of vector spaces for morphologically richer languages.


\section{Conclusion}

We have presented a novel \textsc{Attract-Repel} method for injecting linguistic constraints into word vector space representations. The procedure \emph{semantically specialises} word vectors by jointly injecting mono- \emph{and} cross-lingual synonymy and antonymy constraints, creating unified cross-lingual vector spaces which achieve state-of-the-art performance on the well-established SimLex-999 dataset and its multilingual variants. Next, we have shown that \textsc{Attract-Repel} can induce high-quality vectors for lower-resource languages by tying them into bilingual vector spaces with high-resource ones. We also demonstrated that the substantial gains in intrinsic evaluation translate to gains in the downstream task of dialogue state tracking (DST), for which we release two novel non-English datasets (in German and Italian). Finally, we have shown that our semantically rich cross-lingual vectors facilitate language transfer in DST, providing an effective method for bootstrapping belief tracking models for new languages.  

\subsection{Further Work}

Our results, especially with DST, emphasise the need for improving vector space models for morphologically rich languages. Moreover, our intrinsic and task-based experiments exposed the discrepancies between the conclusions that can be drawn from these two types of evaluation. We consider these to be major directions for future work.

\section*{Acknowledgements}

The authors would like to thank Anders Johannsen for his help with extracting BabelNet constraints. We would also like to thank our action editor Sebastian Pad\'{o} and the anonymous TACL reviewers for their constructive feedback. Ivan Vuli\'{c}, Roi Reichart and Anna Korhonen are supported by the ERC Consolidator Grant LEXICAL (number 648909). Roi Reichart is also supported by the Intel-ICRI grant: Hybrid Models for Minimally Supervised Information Extraction from Conversations.



\bibliography{tacl2016}
\bibliographystyle{acl2012}


\end{document}